\documentclass[11pt]{article}

\usepackage[utf8]{inputenc}
\usepackage{icml2017,palatino} 
\usepackage{times}
\usepackage{graphicx} 
\usepackage{float}
\usepackage{subfig} 
\usepackage{enumitem}
\usepackage{natbib}

\usepackage{ulem}
\usepackage{verbatim}

\usepackage{algorithm}
\usepackage[noend]{algpseudocode}

\usepackage{amsmath}
\usepackage{amssymb}

\usepackage{hyperref}
\hypersetup{
    colorlinks,
    linkcolor={red!50!black},
    citecolor={blue!50!black},
    urlcolor={blue!80!black}
}

\title{Independently Controllable Features}

\author{
Emmanuel Bengio \\
McGill University\\
\texttt{ebengi@cs.mcgill.ca} \\
\And
Valentin Thomas \\
École Polytechnique Fédérale de Lausanne\\
\texttt{valentin.thomas@epfl.ch} \\
\AND
Joelle Pineau \\
McGill University \\
\texttt{jpineau@cs.mcgill.ca} \\
\And
Doina Precup \\
McGill University \\
\texttt{dprecup@cs.mcgill.ca} \\
\And
Yoshua Bengio\\
CIFAR Senior Fellow, Université de Montréal \\
\texttt{yoshua.bengio@umontreal.ca} \\
}

\begin{document} 

\maketitle









\begin{abstract} 
Finding features that disentangle the different causes of variation in real data is a difficult task, that has nonetheless received considerable attention in static domains like natural images. Interactive environments, in which an agent can deliberately take actions, offer an opportunity to tackle this task better, because the agent can experiment with different actions and observe their effects. We introduce the idea that in interactive environments, latent factors that control the variation in observed data can be identified by figuring out what the agent can control. We propose a naive method to find factors that explain or measure the effect of the actions of a learner, and test it in illustrative experiments.
\end{abstract} 

\acknowledgements{The authors gratefully acknowledge financial support for this work by the Samsung Advanced Institute of Technology (SAIT), the Canadian Institute For Advanced Research (CIFAR), as well as the Natural Sciences and Engineering Research Council of Canada (NSERC).}  

\startmain

\section{Introduction}
Whether in static or dynamic environments, decision making for real world problems is often confronted with the hard challenge of finding a ``good'' representation of the problem. 
In the context of supervised or semi-supervised learning, it has been argued~\citep{Bengio-2009-book} that good representations separate out underlying explanatory factors, which may be causes of the observed data.
In such problems, feature learning often involves  mechanisms such as  autoencoders \citep{hinton2006reducing}, which find latent features that explain the observed data. In interactive environments, the temporal dependency between successive observations creates a new opportunity to notice structure in data which may not be apparent using only observational studies. The need to experiment in order to discover causal structures has already been well explored in psychology (e.g.~\cite{gopnik}). In reinforcement learning, several approaches explore  mechanisms that push the internal representations of  learned models to be ``good'' in the sense that they provide better control (see Sec.~\ref{sec:related}).

We propose and explore a more direct mechanism, which explicitly links an agent's control over its environment with its internal feature representations. Specifically, we hypothesize that some of the factors explaining  variations in the data correspond to aspects of the world which can be controlled by the agent. For example, an object could be pushed around or picked up independently of others. In such a case, our approach aims to extract object features  from the raw data while learning a policy that controls precisely these features of the data. 
In Sec.~\ref{sec:icf} we explain this mechanism and show experimental results in its simplest instantiation. In Sec.~\ref{sec:general} we discuss how this principle could be applied more generally, and what are the research challenges that emerge.


\section{Independently controllable features}
\label{sec:icf}

To make the above intuitions concrete, assume that there are factors of variation underlying the  observations coming from an interactive environment that are ``independently controllable''. That is, for each of these factors of variation, there exists a policy which will modify that factor only, and not the others. 
For example, the object behind a set of pixels could be acted on independently from other objects, which would explain variations in its pose and scale when we move it around. The object in this case is a ``factor of variation".
What makes discovering and mapping such factors into features tricky is that the factors are not explicitly observed. Our goal is to learn these factors, which we call~\textbf{independently controllable features}, along with policies that control them.
While these may seem strong assumptions about the nature of the environment, our point of view is that they are similar to regularizers meant to make a difficult learning problem better constrained.

There are many possible ways to express the desire to learn independently controllable features as an objective. Section \ref{sec:pol-sel} proposes such an objective for a simple scenario. Section \ref{sec:exp-emmanuel} illustrates the effect of this objective when all the features of the environment are simple and controllable by the agent. Section \ref{sec:exp-valentin} explores a slightly harder scenario in which there is redundancy and policies are learned through a reinforcement learning algorithm.

\subsection{Autoencoders}
Our approach builds on the familiar framework of 
autoencoders~\citep{hinton2006reducing}, which are defined as a pair of function approximators $f,g$ with parameters $\theta$ such that $f:X\to H$ maps the input space to some \textit{latent space} $H$, and $g:H\to X$ maps back to the input space $X \subset \mathbb{R}^d$. Autoencoders are trained to minimize the discrepancy between $x$ and $g(f(x))$, a.k.a. the reconstruction error, e.g.,:
$$\min_\theta \tfrac{1}{2} \|x-g(f(x))\|_2^2$$
We call $f(x)=h\in H \subset \mathbb{R}^n$ the latent feature representation of $x$, with $n$ features.

It is common to assume that  $n \ll d$. This causes $f$ and $g$ to perform dimensionality reduction of $X$, i.e. \textit{compression}, since there is a dimension bottleneck through which information about the input data must pass. Often, this bottleneck forces the optimization procedure to uncover principal factors of variation of the data on which they are trained. However, this does not necessarily imply that the different
dimensions of $h=f(x)$ are individually meaningful. In fact, note that for any bijective function $r$, we
could obtain the same reconstruction error by
replacing $f$ by $r \circ f$ and $g$ by $r^{-1} \circ g$, so we should not expect any form of disentangling
of the factors of variation unless some additional
constraints or penalties are imposed on $h$. This
motivates the approach we are about to present. Specifically, we will look for
policies which can separately influence one of the dimensions of $h$, and we will prefer representations which make such policies possible.

\subsection{Policy Selectivity}
\label{sec:pol-sel}


Consider the following simple scenario: we train an autoencoder $f,g$ producing $n$  latent features,  $f_k,k=1,\dots n$. In tandem with these features we train $n$ policies, denoted $\pi_k$. Autoencoders can learn relatively arbitrary feature representations, but we would like
these features to correspond to controllable
factors in the learner's environment. Specifically, we would like policy $\pi_k$ to cause a change only in $f_k$ and not in any other features. We think of $f_k$ and $\pi_k$ as a feature-policy pair.

In order to quantify the change in $f_k$ when actions are taken according to $\pi_k$, we define the  \emph{selectivity} of a feature as:
\begin{equation}
    sel(s,a,k) = \mathbb{E}_{s'\sim \mathcal{P}_{ss'}^a} \left[ \frac{|f_k(s')-f_k(s)|}{\sum_{k'}|f_{k'}(s')-f_{k'}(s)|} \right] \label{eq:sel}
\end{equation}
where $s$,$s'$ are successive  raw state representations (e.g. pixels), $a$ is the action, $\mathcal{P}_{ss'}^a$ is the environment's transition distribution from $s$ to $s'$ under action $a$. The normalization by the change in all features means that the selectivity of $f_k$ is maximal when \textit{only that single feature} changes as a result of some action. 

By having an objective that maximizes selectivity \textit{and} minimizes the autoencoder objective, we can ensure that the features  learned can both reconstruct the data and  recover 
independently controllable factors. Hence, we define the following objective, which can be minimized via stochastic gradient descent:
\begin{equation}
     \underbrace{\mathbb{E}_s [\tfrac{1}{2} || s - g(f(s))||^2_2]}_{\text{reconstruction error}}\ -\ \lambda \underbrace{\sum_k \mathbb{E}_{s}[\sum_a \pi_k(a|s) \log sel(s, a, k)   ]}_{\text{disentanglement objective}}  \label{eq:full-objective}
\end{equation}
Here one can think of $\log sel(s,a,k)$ as the reward signal $R_k(s,a)$ of a control problem, and the expected reward $E_{a\sim \pi_k}[R_k]$ is maximized by finding the optimal set of policies $\pi_k$.




Note that it is also possible to have \textit{directed} selectivity: by not taking the absolute value of the numerator of \eqref{eq:sel} (and replacing $\log sel$ with $\log (1+sel)$ in \eqref{eq:full-objective}), the policies must learn to increase the learned latent feature rather than simply change it. This may be useful if the policy to gradually increase a feature is distinct from the policy that decreases it.




\subsection{A first toy problem}
\label{sec:exp-emmanuel}

\begin{figure}[H]
\centering
\includegraphics[width=0.5\columnwidth,trim={0 30px 0 0},clip]{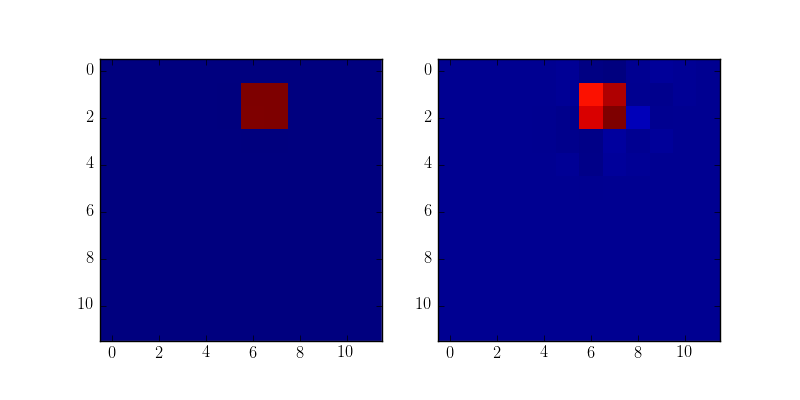}
\includegraphics[width=0.2\columnwidth,trim={75px 0 0 0},clip]{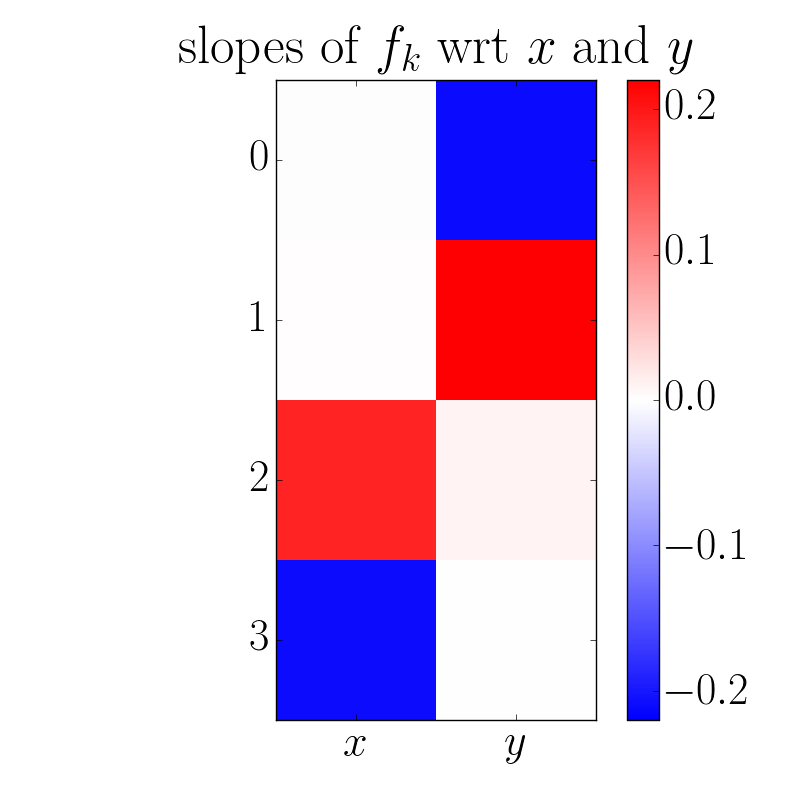}
\includegraphics[width=0.23\columnwidth]{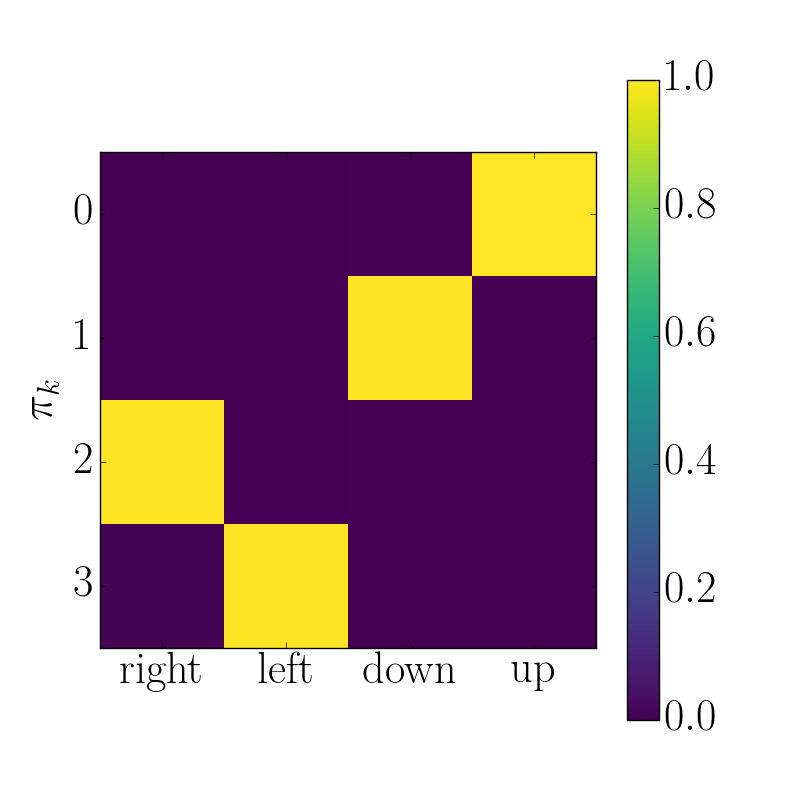}

\hspace*{0.1cm}(a)\hspace*{3.5cm}(b)\hspace*{3.9cm}(c)\hspace*{3.7cm}(d)
\caption{(a) \& (b) A simple environment with 4 actions that push a square left, right, up or down. (a) is an example ground truth, (b) is the reconstruction of the model trained with selectivity. (c) The slope of a linear regression of the true features (the real $x$ and  $y$ position of the agent) as a function of each latent feature. White is no correlation, blue and red indicate strong negative or positive slopes respectively. We can see that features 0 and 1 recover $y$ and features 2 and 3 recover $x$. (d) Representation of the learned policies. Each row is a policy $\pi_k$, each column corresponds to an action (left/right/up/down). Each cell $(k,i)$ represents the probability of action $i$ in policy $\pi_k$;  We can see that features 0 and 1 correspond to going down and up ($-y$/$+y$) and features 2 and 3 correspond to going right and left ($+x$/$-x$).}
\label{fig:squares-env-result}
\end{figure}

Consider the simple environment described in Figure \ref{fig:squares-env-result}(a): the agent sees a $2\times 2$ square of adjacent cells
in the environment, and has 4 actions that move it up, down, left or right. An autoencoder with directed selectivity  (see Figure \ref{fig:squares-env-result}(c,d)) learns latent features that  map to the $(x,y)$ position of the square in the input space, without ever having explicitly access to these values, and while reconstructing the input properly.
An autoencoder without selectivity also reconstructs the input properly but without learning these two latent $(x,y)$ features explicitly. 

In this setting $f$, $g$ and $\pi$ share some of their parameters. We use the following architecture: $f$ has two $16\times 3 \times 3$ ReLU convolutional layers, followed by a fully connected ReLU layer of 32 units, and a $\tanh$ layer of $n=4$ features; $g$ is the transpose architecture of $f$; $\pi_k$ is a softmax policy over 4 actions, computed from the output of the ReLU fully connected layer.



\subsection{A slightly harder toy problem}
\label{sec:exp-valentin}
In the next experiments, we aim to generalize the model above in a slightly more complex environment. 
Instead of having $f$ and $\pi_k$ parametrized by the same parameters, we now introduce a different set of parameters for each policy and for the encoder, so that each policy can be learned separately.

Additionally, we use a richer action space, in which we added a ``move down and to the right" action as well as an ``increase/decrease color of  square" action. Note also that the first two actions ("move down") are redundant.

Thus, in this setting, $f, g$ and $\pi_k$ are all parametrized by different variables, as follows:
\begin{itemize}[nosep, topsep=-1ex]
\item $f(s) = f(s; W_f)$ a neural net with a $\tanh$ output activation 
\item $g(h) = g(h; W_g)$ a neural net with a ReLU output activation
\item $\pi_k (a|s) = \pi(a|s ; \theta_k)$ so that $\pi_k( \cdot | s)  = \text{softmax}(\theta_k \cdot s)$
\end{itemize}

\begin{algorithm}                      
\caption{Training an autoencoder with disentangled factors}          
\label{alg1}                           
\begin{algorithmic}[1]                   
    \For{$t=1..T$}
        \State{Sample $s$}
        \State{$W_f \gets W_f - \eta_{f} \nabla_{W_f} [ \tfrac{1}{2} ||s -g (f(s))||^2_2]$} 
        \State{$W_g \gets W_g -  \eta_{g} \nabla_{W_g} [ \tfrac{1}{2} ||s -g (f(s))||^2_2]$} 
        \For{$k=1..K$}
            \State{$W_f \gets W_f + \eta_{f} \ \lambda\  \nabla_{W_f} \mathbb{E}_{a \sim \pi_k(\cdot | s)} [ \log  sel(s,a,k) ]$} 
            \State{$\theta_k \gets \theta_k + \eta_{k} \ \lambda \ \nabla_{\theta_k}  \mathbb{E}_{a \sim \pi_k(\cdot | s)} [\log  sel(s,a,k) ]$}
        \EndFor
    \EndFor
\end{algorithmic}
\end{algorithm}

The gradients on lines 3 and 4 are computed exactly via backpropagation. In our experiments, gradients on lines 6 and 7 are also computed by backpropagation, but 
in a more general case in which the environment also provides a reward $r_t$ (so the total reward to maximize becomes $r_t + \log sel_k$), they can be estimated via Monte-Carlo. Note that, in any case, $\nabla_{\theta_k}  \mathbb{E}_{a \sim \pi_k(\cdot | s)} [\log  sel(s,a,k) ]$ can be computed
with the REINFORCE \citep{williams1992simple} estimator: 
$$\nabla_{\theta_k} \mathbb{E}_{a \sim \pi_k(\cdot | s)}[ \log sel(s, a, k) ] = \mathbb{E}_{a \sim \pi_k(\cdot | s)}[ \log sel(s, a, k) \cdot \nabla_{\theta_k} \log \pi_k(a|s)]$$

\begin{figure}[H]
\centering
\subfloat[]{\includegraphics[width=.3\linewidth]{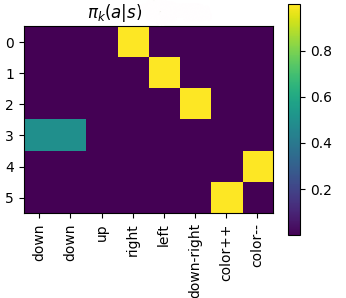}}
\hfill
\subfloat[]{\includegraphics[width=.3\linewidth]{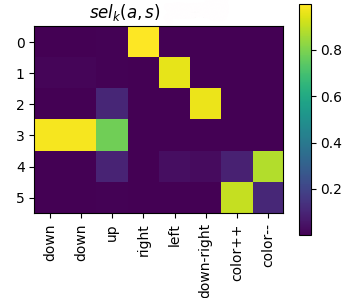}}
\hfill
\subfloat[]{\includegraphics[width=.3\linewidth]{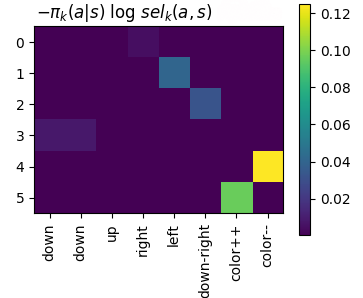}}
\caption{Policy $\pi_k$, selectivity $sel_k$ and objective $-\pi_k  \log sel_k$ for a random $s$ and feature/policy $k$ after optimizing (2). On the $y$-axis is the index $k$ of the feature, and on the $x$-axis, the $8$ different actions that the agent can select. Initially, $\pi_k$ was initialized to be uniform.
(a): As in the previous toy problem, each $\pi_k$ quickly concentrates on a specific action. 
(b): However, in this example, the encoder is also simultaneously shaped so that each feature $f_k(s)$ reacts to one specific action with maximal selectivity (L6 of algorithm \ref{alg1})  (c): $- \pi_k \log sel_k$. The autoencoder was able to learn independently controllable features.
Moreover, in (a), we observe that, as the two first actions are redundant, they have the exact same selectivity and therefore $\pi_3$, the policy associated to the "move down" feature, can either choose the first or second action. }
\end{figure}
\section{Scaling to general environments: controllability and the binding problem}
\label{sec:general}


In the previous section we used problems in which  the environment is made of a static set of objects. In this case,   if the objective posited in section \ref{sec:pol-sel} is learned correctly, we can assume that feature $k$ of the representation can unambiguously refer to some controllable property of some specific object in the environment. For example, the agent's world might contain only a red circle 
and a green rectangle, which are only affected bu the actions of the agent (they do not move on their own) and we only change the positions and colours of these objects from one trial 
to the next. Hence, a specific feature $f_k$ can learn to unambiguously refer to the position or the colour of one of these two objects.

In reality, environments are stochastic, and the set of objects  in a given scene is drawn from some distribution. The number of objects may vary and their types may be different. 
It then becomes less obvious how feature $k$ could refer in a clear way to some feature of one of the objects in a particular scene. 
If  we have \textit{instances} of objects of different types, some addressing or naming scheme is required to refer to the particular objects (instances) present in the scene, so as to match the policy with a particular attribute of a particular object to selectively modify. In our simple environments, this task was trivial because we could simply use the integer $k$ to achieve this coordination between the policy $\pi_k$ and the representation elements (feature $f_k(s)$). In the more general case, this is not possible, which gives rise to a representational problem.

This is connected to the binding problem in neuro-cognitive science: how to represent a set of objects, each  having different attributes, so that we don't confuse, for example, the set $\{$red circle, blue square$\}$ with $\{$red square, blue circle$\}$. It is not enough to have a red-detector feature and a blue-detector feature, a square-detector feature and a circle-detector feature. The binding problem has seen some attention in the representation learning literature \citep{minin2012complex,greff2016tagger}, 
 but still remains mostly unsolved. 
Jointly considering this problem and larning controllable features may prove fruitful.

These ideas may also lead to interesting ways of performing exploration.
How do humans choose with which object to play? We are attracted to objects which we do not know yet (i.e., 
if and how we can control them). The RL exploration process could be driven by a notion of controllability, predicting the interestingness of objects in a scene and choosing features and associated policies with which to attempt control them. Such ideas have only been explored briefly in the literature, e.g.~\cite{ratitch}
%



\section{Discussion}
\label{sec:related}
There is a large body of work on learning features in RL focusing on indirectly learning good internal representations. In \cite{jaderberg2016reinforcement}, agents learn off-policy to control their pixel inputs, forcing them to learn features that, intuitively, help  control the environment (at least at the pixel level). \cite{oh2015action} propose models that learn to predict the future, conditioned on action sequences, which push the agent 
to capture temporal features.  There are many more works that go in this direction, such as (deep) successor feature representations \citep{dayan1993improving,kulkarni2016deep} or the options framework \citep{sutton1999between,precup2000temporal} when used in conjunction with neural networks \citep{bacon2016option}.

Our approach is similar in spirit to the Horde architecture~\citep{sutton2009}. In that scenarion, agents learn policies that maximize specific inputs. The predictions for all these policies then become features for the agent. Our objective is defined specifically in the context of autoencoders. Unlike recent work on the predictron~\citep{silver2017}, our approach is not focused on solving a planning task, and the goal is simply to learn how the world works.


\vspace*{-10pt}

\bibliography{main}
\bibliographystyle{icml2017}

\end{document}